\documentclass[10pt,twocolumn,letterpaper]{article}

\usepackage{cvpr}              
\usepackage{enumitem}
\usepackage{amsthm,amsmath,amssymb,mathrsfs}
\usepackage{nicefrac}       
\usepackage{booktabs}
\usepackage{multirow}
\usepackage[normalem]{ulem}
\useunder{\uline}{\ul}{}
\usepackage{enumitem}

\usepackage[dvipsnames]{xcolor}

\definecolor{cvprblue}{rgb}{0.21,0.49,0.74}
\usepackage[pagebackref,breaklinks,colorlinks,citecolor=cvprblue]{hyperref}

\def\paperID{12139} 
\def\confName{CVPR}
\def\confYear{2024}

\title{Learning Dynamic Tetrahedra for High-Quality Talking Head Synthesis}

\author{
\large {Zicheng Zhang$^1$}
\quad {Ruobing Zheng$^2$}
\quad {Ziwen Liu$^1$}
\quad {Congying Han$^1$\thanks{Corresponding author}}
\quad {Tianqi Li$^2$} \\
\quad {Meng Wang$^2$}
\quad {Tiande Guo$^1$}
\quad {Jingdong Chen$^2$}
\quad {Bonan Li$^1$}
\quad {Ming Yang$^2$}\\
\normalsize \textsuperscript{\rm 1}University of Chinese Academy of Sciences
\quad
\textsuperscript{\rm 2}Ant Group
}

\begin{document}
\maketitle
\begin{abstract}
Recent works in implicit representations, such as Neural Radiance Fields (NeRF), have advanced the generation of realistic and animatable head avatars from video sequences. These implicit methods are still confronted by visual artifacts and jitters, since the lack of explicit geometric constraints poses a fundamental challenge in accurately modeling complex facial deformations. In this paper, we introduce Dynamic Tetrahedra (DynTet), a novel hybrid representation that encodes explicit dynamic meshes by neural networks to ensure geometric consistency across various motions and viewpoints. DynTet is parameterized by the coordinate-based networks which learn signed distance, deformation, and material texture, anchoring the training data into a predefined tetrahedra grid. Leveraging Marching Tetrahedra, DynTet efficiently decodes textured meshes with a consistent topology, enabling fast rendering through a differentiable rasterizer and supervision via a pixel loss. To enhance training efficiency, we incorporate classical 3D Morphable Models to facilitate geometry learning and define a canonical space for simplifying texture learning. These advantages are readily achievable owing to the effective geometric representation employed in DynTet. Compared with prior works, DynTet demonstrates significant improvements in fidelity, lip synchronization, and real-time performance according to various metrics. Beyond producing stable and visually appealing synthesis videos, our method also outputs the dynamic meshes which is promising to enable many emerging applications. Code is available at \url{https://github.com/zhangzc21/DynTet}.
\end{abstract}
\section{Introduction}
\label{sec:intro}

Talking head synthesis is a long-standing task with a wide range of applications, such as digital humans, metaverse and filmmaking. The set up of this task can be roughly categorized to two lines: 1) learning from a large-scale dataset to drive arbitrary portrait images by the motion signal~\cite{chen2019hierarchical, prajwal2020wav2lip, thies2020nvp, zhou2020makelttalk, zhou2021pcavs, lu2021lsp, zhang2021facial,Ma2023OTAvatarOT}; and 2) building a personalized animatable head model from a several-minute video of a specific person~\cite{Zheng2021IMA,Thies2016Face2FaceRF,Yi2020AudiodrivenTF,guo2021ad, liu2022semantic, shen2022dfrf, yao2022dfa,tang2022rad,Li2023EfficientRN}.  We concentrate on the latter one, since it generally delivers high-quality synthesis results with intricate details and 3D naturalness, suitable for professional scenarios.

Building such an exquisite talking head avatar from video data poses challenges in faithful appearance, motion control, as well as low running cost. 
One line of methods~\cite{Kim2018DeepVP,Yi2020AudiodrivenTF,Garrido2016ReconstructionOP,Thies2016Face2FaceRF,Geng20193DGF,Karras2017AudiodrivenFA,Fan2021FaceFormerS3} explicitly rely on 3D Morphable Models (3DMM)~\cite{Blanz1999AMM}  to reconstruct and animate human faces by estimating the person-specific parameters. While these methods allow for efficient rendering and  dynamic deformation, the fixed face topology makes them often fall short in generating characteristic details,~\textit{e.g.}, hairstyle, glasses, and inner mouth. Recently, neural implicit representations, especially
Neural Radiance Fields (NeRF)~\cite{mildenhall2021nerf}, provide a new way to realize faithful generation. Some seminal work~\cite{Gafni2020DynamicNR,guo2021ad, liu2022semantic, shen2022dfrf, yao2022dfa}  learn a direct mapping from the control signal to the talking head, meanwhile the efficient neural representations like  voxel grid~\cite{sun2022direct,yu2021plenoxels,Liu2022DeVRFFD} and hash encoding~\cite{muller2022instant} have been introduced to improve training and inference speed.

Despite the expressive capabilities, these implicit methods still need to improve many subtle issues in realism,~\textit{e.g.}, head jitters, motionless mouths, and occasional artifacts. 
Researchers have discussed these issues from various aspects, \textit{e.g.}, introducing a canonical space for easier appearance learning~\cite{Park2020NerfiesDN,Athar2022RigNeRFFC}, developing compact models for more efficient training~\cite{tang2022rad,Li2023EfficientRN}, 
and using expressive driving conditions for better control \cite{yao2022dfa,ye2023geneface},  against the naive baseline of NeRF. 
Essentially, the implicit definition for 3D objects complicates the analytical alternation of the underlying object geometry, leading to ineffective disentanglement of static appearance and motion from dynamic data, as opposed to the explicit meshes and vertex displacements for 3DMM.
In viewing of this, it is appealing to incorporate the expressivity of implicit methods with an effective geometric control to take advantage of both lines of works.

We introduce Dynamic Tetrahedra (DynTet), a novel \textit{hybrid approach} that encodes dynamic meshes within neural networks to assist explicit deformation.  
In essence, DynTet employs neural networks to predict attributes of underlying surfaces, from which  explicit meshes can be extracted to fast render images with a differentiable rasterizer. On one hand, distinguished from implicit methods, the explicit geometry enables DynTet to learn a consistent 3D model across frames and directly express deformation as vertex displacement, thus DynTet is convenient to model the dynamics of talking head.  On the other hand, unlike 3DMM with a preset topology, DynTet end-to-end learns personalized meshes and texture suitable for the given video.

Technically, the proposed DynTex is inspired by recent advancements in tetrahedral techniques~\cite{Gao2020LearningDT}, originally developed for 3D reconstruction~\cite{Munkberg2021ExtractingT3} and synthesis~\cite{Gao2022GET3DAG,Shen2021DeepMT}. 
The key insight is a parameterized tetrahedral grid: Coordinate-based networks are used to learn the signed distance field~(SDF) and refinement for the grid, then the meshes are subsequently decoded through the Marching Tetrahedra algorithm.
Since these prior methods~\cite{Gao2020LearningDT,Shen2021DeepMT,Munkberg2021ExtractingT3,Gao2022GET3DAG} primarily work for static scenes, they can hardly render images with deformation, nor keep the mesh topology under different conditions~\cite{Gao2022GET3DAG}. 
This also largely prevents the parallelized rendering and training processes.
In contrast, we redesign the framework that exclusively determines the mesh topology with SDF, while a new branch controls the geometric variations with deformation signals. We also estimate the elastic score of each vertex to specify the rigid (\textit{e.g.}, forehead) and non-rigid regions (\textit{e.g.}, mouth) of the head, so as to achieve precise deformations in local regions while maintaining stability in other parts. These designs ensure the topological consistency and expressivity across all decoded meshes, enabling a customized dynamic mesh beyond 3DMM.
To improve training efficiency, we introduce geometry losses that supervise shape and motion using the 3DMM priors. This replenishes the limited depth information available in frontal talking videos. Moreover, we establish an interpretable canonical space for the dynamic mesh, reducing the complexity of texture learning. In summary, our contributions include: 
\begin{itemize}[leftmargin=0em, itemindent=3em]
    \item[(\textbf{1}).] We propose DynTet, a novel hybrid representation that encodes dynamic head meshes in neural networks, where the explicit geometry delineated by tetrahedra facilitates  appearance and motion learning.
    \item[(\textbf{2}).] This is the first work that successfully extends the static tetrahedral representation to dynamic head avatars by a new elaborated architecture, a canonical space, and 3DMM guidance for modeling dynamic meshes. 
    \item[(\textbf{3}).] DynTet presents evident advantages in terms of fidelity, lip-sync precision, stability and runtime by thorough evaluation compared with prior works.
\end{itemize}

The learned dynamic head meshes are promising together with existing 3D assets or AR/VR techniques for emerging applications such as human avatar and the metaverse, which may inspire further study on the hybrid representation for dynamic 3D objects.

\section{Related Work}

\paragraph{Talking head synthesis. }
Most existing methods for talking head synthesis can be classified into three categories in terms of the modeling approaches.
\textit{2D-based methods}~\cite{prajwal2020wav2lip,zhou2020makelttalk,yu2020multimodal,zhou2019talking,yin2022styleheat} utilized generative models~\cite{goodfellow2020generative,isola2017image,Karras2019AnalyzingAI} as renderers to produce photorealistic portraits. While, these methods often fall short in achieving 3D naturalness and consistent pose control due to the absence of an explicit 3D model.
 \textit{3DMM-based methods}~\cite{Kim2018DeepVP,Yi2020AudiodrivenTF,Garrido2016ReconstructionOP,Thies2016Face2FaceRF,Geng20193DGF,Karras2017AudiodrivenFA,Fan2021FaceFormerS3} leveraged 3D face knowledge to drive facial expression, resulting in quite natural talking style. As a trade-off, due to the lack of complete head topology they cannot guarantee consistency or fidelity beyond the facial region such as hair and inner mouth.
\textit{Neural head avatar}~\cite{Grassal2021NeuralHA,Zheng2021IMA,Zheng2022PointAvatarDP} has emerged as an attractive way for automatically creating 3D facial models. While earlier approaches~\cite{Yenamandra2020i3DMMDI,Ramon2021H3DNetFH} required detailed scanning data, NeRFs~\cite{mildenhall2021nerf,Park2020NerfiesDN,chan2022efficient} offer a solution to learn models from video clips, driven by the factors like 3DMM coefficients~\cite{Gafni2020DynamicNR}, audio feature~\cite{guo2021ad}, or facial landmarks~\cite{ye2023geneface}. Several recent works have enhanced control effects  ~\cite{Athar2022RigNeRFFC,liu2022semantic,yao2022dfa}, system efficiency~\cite{tang2022rad,Li2023EfficientRN} and few-shot training~\cite{shen2022dfrf,Li2023GeneralizableON}. Despite great progress, neural avatars still face challenges in realism and plausible motion, 
not to mention that NeRF rendering is computationally expensive when generating high-resolution images. 

\vspace{-3em}
\paragraph{Neural 3D representation.}
Neural implicit functions are emerging as an effective representation of 3D object~\cite{Park2019DeepSDFLC,Mescheder2018OccupancyNL,Liu2020NeuralSV,mildenhall2021nerf,Sitzmann2019SceneRN} using  coordinate-based  MLPs~\cite{Tancik2020FourierFL,Sitzmann2020ImplicitNR,muller2022instant}. 
While prior methods~\cite{Park2019DeepSDFLC,Mescheder2018OccupancyNL} required 3D supervision, several recent works~\cite{Liu2019LearningTI,Liao2018DeepMC,Niemeyer2019DifferentiableVR,mildenhall2021nerf} demonstrated differentiable rendering for training directly from images. 
Meanwhile, NeRF~\cite{mildenhall2021nerf} and followups~\cite{sun2022direct,yu2021plenoxels,muller2022instant,Zhang2023TransformingRF} performed volume rendering~\cite{Kajiya1984RayTV} on a continuous field for density and color, achieving impressive results for multi-view synthesis. As density is ambiguous to depict geometric details~\cite{Wang2021NeuSLN}, they cannot explicitly edit shape or extract a high-quality surface with Marching Cubes~\cite{Lorensen1987MarchingCA}, which makes it challenging in formulating deformation for dynamic scenes.
Recent works~\cite{Gao2020LearningDT,Shen2021DeepMT} proposed to convert deformable tetrahedral grid into surface meshes via the differentiable Marching Tetrahedral algorithm~\cite{marchingtet}, where the SDF values are implicitly learned. The following methods~\cite{Munkberg2021ExtractingT3,Gao2022GET3DAG} further extended to jointly learn geometry and texture from image data. They can render photorealistic images competitive with NeRF in real-time.  
However, as these methods only work for static scenes, they cannot learn dynamics for modeling head avatars. We present an original effort to address this issue to learn dynamic tetrahedral meshes.

\section{Preliminaries}\label{sec:Preliminaries}
We provide some background concepts and notations. At first, we define operations between sets as element-wise operations, and functions are mapped element-wise as well.

\noindent\textbf{Tetrahedral meshes} have been studied in deep learning~\cite{Gao2020LearningDT,Shen2021DeepMT} as a hybrid representation for 3D shape modeling. Consider an object lying in a unit cube, a tetrahedral grid denoted as ($\mathbf{V}_{tet}$, $\mathbf{T}_{tet}$) is pre-defined to tetrahedralize the cube into  $r^3$ resolution. Each tetrahedron $T_k \in \mathbf{T}_{tet}$ is defined by four vertices $\{\mathbf{v}_{a_k}, \mathbf{v}_{b_k}, \mathbf{v}_{c_k}, \mathbf{v}_{d_k}\}$, where $K$ is the total number of tetrahedra. 
Additionally, each vertex $\mathbf{v}_{i}$ contains a learnable SDF value $s_i \in \mathbb{R}$ to express the distance away from the underlying surface, and a small offset $\Delta \mathbf{v}_{i}  \in [-\frac{1}{r},\frac{1}{r}]^3$ from its initial coordinates to refine the grid as $\mathbf{v}^{\prime}_i=\mathbf{v}_i + \Delta \mathbf{v}_{i}$.
We let $\mathcal{S} = \{s_i\}^{K}_{i=1}$ and $\Delta \mathbf{V}  = \{\Delta \mathbf{v}_i\}^{K}_{i=1}$. 
Notably, the surface shape within the tetrahedral grid is primarily determined by the SDF values $\mathcal{S}$ as $\Delta \mathbf{V}$ is confined within the resolution boundaries.

\begin{figure*}[t]
	\centering
	\includegraphics[width=1\linewidth]{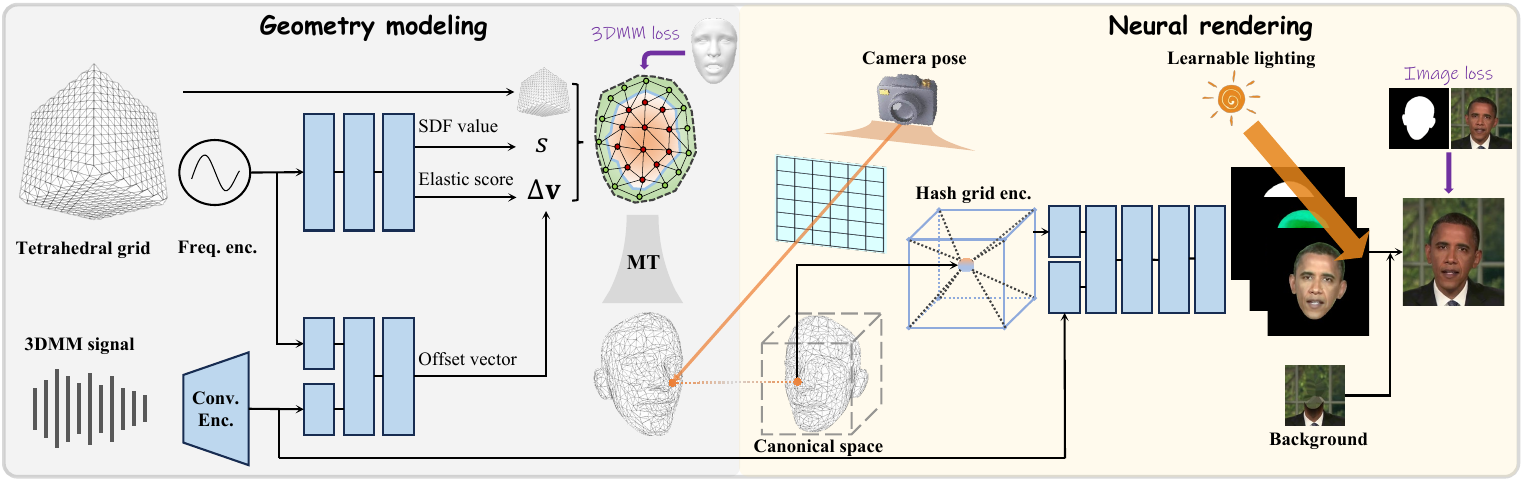}
	\caption{Illustration of the proposed DynTet for modeling a talking head. Left panel: The top branch predicts topology-related information, including SDF values $s$ and elastic scores. The bottom branch, conditioned by talking signals, predicts the offset vectors scaled by the elastic scores to yield deformation vectors $\Delta \mathbf{v}$ for driving the tetrahedral grid. Then, the Marching Tetrahedra (\textit{MT}) algorithm decodes the meshes. Right panel: The pixel coordinates are projected into a canonical space. Then the physically-based materials and lighting are sampled for rendering images.   }
	\label{fig:overview}
	\vspace{-3mm}
\end{figure*}

\noindent\textbf{Marching Tetrahedra} (\textit{MT}) is an iso-surface extraction algorithm~\cite{marchingtet} to generate triangular meshes from tetrahedral meshes.
Given SDF values $\{s_a, s_b, s_c, s_d\}$ of a tetrahedron, \textit{MT} determines the surface topology inside the tetrahedron based on the signs of these values, where the $2^4$ configurations in total fall into 3 unique cases after considering rotation symmetry~\cite{Shen2021DeepMT}. A simplified 2D example can be found in Figure \ref{fig:overview}. Once the topology is identified, a new vertex denoted as $\mathbf{v}_{ab}$ by example, is located at the zero crossings of linear interpolation along the tetrahedral edges:
\begin{equation}\label{eq:interp}
\setlength\abovedisplayskip{0.5em}
\setlength\belowdisplayskip{0.5em}
    \mathrm{lerp}(\mathbf{v}_a,\mathbf{v}_b,s_a,s_b) = \frac{s_a \cdot \mathbf{v}_{a} - s_b \cdot \mathbf{v}_{b}}{s_a-s_b},
\end{equation}
the same for other vertices. Note that this equation is evaluated only when signs differ ($\text{sign}(s_a)\neq \text{sign}(s_b)$) to prevent singularity. 
Leveraging \textit{MT} on the deformed tetrahedral grid ($\mathbf{V}_{tet} + \Delta \mathbf{V} $, $\mathbf{T}_{tet}$), the triangular surface of the encoded object can be acquired in a differentiable manner. We omit constant symbols and represent this process as
\begin{equation}\label{eq:mt}
\setlength\abovedisplayskip{0.5em}
\setlength\belowdisplayskip{0.5em}
	\mathbf{V}_{tri}, \mathbf{T}_{tri} = \mathrm{MarTet}(\mathcal{S}, \Delta \mathbf{V}),
\end{equation} 
where $\mathbf{V}_{tri}$ and $\mathbf{T}_{tri}$ are the sets of vertices and  connectivity of the triangular meshes, respectively.

\noindent\textbf{3D Morphable Models} provide generic parametric models~\cite{Blanz1999AMM} for synthesizing face shapes, usually expressed as
\begin{equation}\label{eq:3dmm}
\setlength\abovedisplayskip{0.5em}
\setlength\belowdisplayskip{0.5em}
	\mathbf{V}_{3dmm}=\bar{\mathbf{V}}_{3dmm}+ \mathbf{U}_{id}\boldsymbol{\gamma}+ \mathbf{U}_{exp}\boldsymbol{\alpha}.
\end{equation}
Here $\bar{\mathbf{V}}_{3dmm}$ denotes a mean shape, $\mathbf{U}_{id}$ and $\mathbf{U}_{exp}$ are the matrices of basis vectors in 
the identity and expression space, respectively. $\boldsymbol{\gamma}$ and $\boldsymbol{\alpha}$ are coefficients for the identity and
expression, allowing for a tractable control over facial variations.  The parameter $\boldsymbol{\alpha}$ has been served as a versatile representation for driving facial deformation~\cite{Egger20193DMF}.

\section{Methodology}
In this section, we introduce a framework called \textit{Dynamic Tetrahedra} (DynTet) to rapidly learn 3D head avatars from short video sequences, and enable real-time rendering of high-quality talking heads. Generally, DynTet upgrades recent tetrahedral representation~\cite{Gao2020LearningDT,Munkberg2021ExtractingT3,Gao2022GET3DAG} tailored for dynamic head modeling. As shown in Figure~\ref{fig:overview}, neural networks are trained to encode head shapes within a tetrahedral grid~\cite{Shen2021DeepMT}, and then the Marching Tetrahedra (\textit{MT})~\cite{marchingtet} decodes the meshes for photorealistic rasterization rendering.  
In the following, we delineate the talking head task through the lens of tetrahedral representation (Sec.\ref{sec:Problem statement}), and elucidate our improvements in the geometry model (Sec.\ref{sec:Geometry model}), rendering process (Sec.\ref{sec:Neural rendering}), and  training losses (Sec.\ref{sec:Loss function}).

\subsection{Problem statement}\label{sec:Problem statement}
Given a sequence of head images $\{\mathbf{I}_i\}_{i=1}^{n}$, we aim to reenact the  specific head driven by motion signals, which typically includes camera parameters $\{\mathbf{P}_{i}\}_{i=1}^{n}$ for rigid motion, and talking signals which we represent using 3DMM expression coefficients $\{\boldsymbol{\alpha}_{i}\}_{i=1}^{n}$ for non-rigid facial expressions. 
\vspace{-1em}
\paragraph{Formulation.} Drawing inspiration from the seminal work of~\cite{Shen2021DeepMT}, which applied tetrahedral meshes for multi-view reconstruction, we propose a basic paradigm for talking head modeling: (1)  A geometry mapping $\boldsymbol{F}$ first predicts $\mathcal{S}$ and $\Delta \mathbf{V}$ from $\boldsymbol{\alpha}$ and $ \mathbf{V}_{tet}$. (2) The outputs from $\boldsymbol{F}$ are processed based on Eq.~\eqref{eq:mt} to obtain the triangular mesh, which are then rasterized given camera pose $\mathbf{P}$ to get the coordinates of image pixels in the model space. (3) An appearance mapping $\boldsymbol{G}$ predicts the materials, which is incorporated with a lighting model $\boldsymbol{L}$ for the physically-based rendering~(PBR)~\cite{Burley2012PhysicallyBasedSA}. We summarize the overall process as
\begin{equation}\label{eq:target}
\setlength\abovedisplayskip{0pt}
          \setlength\belowdisplayskip{1pt}
	\begin{aligned}
		& \min\limits_{\boldsymbol{F},\boldsymbol{G}, \boldsymbol{L}} \ 
		\frac{1}{n} \sum\limits^{n}_{i=1} 
		Loss(\hat{\mathbf{I}}_i, \mathbf{I}_i) \\
		& \text{where} \ \ \hat{\mathbf{I}}_i = \mathrm{Rendering}(\boldsymbol{F}, \boldsymbol{G}, \boldsymbol{L}, \boldsymbol{\alpha}_{i}, \mathbf{P}_{i}).
	\end{aligned}
\end{equation}    
Herein, $\mathrm{Rendering}$ denotes the PBR procedure, and the loss function comprises a set of constraints on the predictions. 
\vspace{-1em}
\paragraph{Challenges.}
Prior works~\cite{Gao2020LearningDT,Shen2021DeepMT,Gao2022GET3DAG} commonly parameterize $\boldsymbol{F}$ using a neural network $f$ to predict SDF values and relative offsets, \textit{i.e.}, $\mathcal{S}, \Delta \mathbf{V} = f(\mathbf{V}_{tet})$, while introducing the talking signal $\boldsymbol{\alpha}$ to $f$ as a condition will bring several problems. First, once SDF values vary along with the conditions, it is difficult to learn the intricate deformation of the facial area with a simple MLP. Second, modern GPUs do not support parallelization for meshes with a variable topology, resulting in inefficient training and running processes. Lastly, previous approaches were designed for $360^{\circ}$ data, while our training data only includes frontal head images. Consequently, this inaccurate estimation of geometry leads to evident artifacts in the rendered videos.

\subsection{Geometry modeling}\label{sec:Geometry model}
To address the mentioned challenges, we initially disentangle topology and geometry from head shapes within the tetrahedra grid, allowing for generating dynamic meshes with a consistent topology. Furthermore, we enhance the tetrahedral attributes by introducing a novel elastic score,  facilitating  precise control over deformation. These improvements are incorporated by modifying the internal structure of the geometry mapping as 
\begin{equation}\label{eq:disentangle}
	\begin{aligned}
		& \boldsymbol{F}:\mathbf{V}_{tet}, \boldsymbol{\alpha} \rightarrow \mathcal{S}, \Delta \mathbf{V}   \\
		&\text{where}\ \  \mathcal{S}, \mathcal{E}= f_{1}(\mathbf{V}_{tet}), \ \Delta \mathbf{V}= \mathcal{E}\cdot f_2(\mathbf{V}_{tet},\boldsymbol{\alpha}).
	\end{aligned}
\end{equation}
Here, $f_1$ and $f_2$ are coordinate-based MLPs, while $\mathcal{E} := [f_{1}(\mathbf{V}_{tet})]_{e}$ represents the set of non-negative elastic scores. 

\vspace{-1em}
\paragraph{Shape representation.} Ideal face models exhibit topological invariance regardless of deformation. While implicit SDF representations present challenges in fulfilling this property, it is clear for explicit meshes like 3DMM~\cite{Blanz1999AMM} to maintain the number of vertices and their relationship.  By employing Eq.~\eqref{eq:interp} within \textit{MT} algorithm, we have
\begin{equation}
	\begin{aligned}
		\mathbf{v}'_{ab} = \underbrace{\mathrm{lerp}(\mathbf{v}_a,\mathbf{v}_b,s_a,s_b)}_{\text{topology and identity}}   + \underbrace{\mathrm{lerp}(\Delta\mathbf{v}_a,\Delta\mathbf{v}_b,s_a,s_b)}_{\text{geometry and expression}}. \\
	\end{aligned}
\end{equation}
Given that vertices $\mathbf{v}_a$ and $\mathbf{v}_b$ are pre-defined in the tetrahedral grid, the first term is just affected by the SDF values. Hence, our design in Eq.\eqref{eq:disentangle}, which segregates SDF from the talking signal, results in a topologically invariant mesh. 
Meanwhile, the second term indicates that the geometry of mesh explicitly relies on the tetrahedral grid offsets. Additionally, a comparison with Eq.~\eqref{eq:3dmm} reveals that \textit{DynTet functions as a quasi-3DMM hybrid model}, enabling additive changes to facial shapes driven by neural networks.

\paragraph{Elastic estimation.} 
Our design necessitates relaxing the range of $\Delta \mathbf{v}_{i}  \in [-\frac{1}{r},\frac{1}{r}]^3$ to allow greater flexibility in geometric variations. However, this relaxation potentially leads to local jitters that significantly impact visual quality. 
We introduce an elastic scoring mechanism for each vertex within the tetrahedral grid to regulate deformation. These scores $\mathcal{E}$ are predicted by the neural network $f_1$ to quantify the non-rigid properties across different regions of the human head. 
 For instance, areas like the forehead and nose exhibit near-rigid behavior with minimal changes during talking, while regions like the mouth and eyes are more flexible and primarily contribute to deformations. 
 In this way, the offset vectors are scaled using the elastic scores to determine the deformable vectors $\Delta \mathbf{V}$. 
 \vspace{-1em}
\paragraph{Architecture details.} As shown in Figure~\ref{fig:overview}, we formulate neural networks $f_1$ and $f_2$ as a composition of regular lightweight MLPs and a frequency positional encoding~\cite{Tancik2020FourierFL} 
\begin{equation}
	\gamma(\mathbf{v})=<\left(\sin \left(2^l \pi \mathbf{v}\right), \cos \left(2^l \pi \mathbf{v}\right)\right)>_{l=0}^{L-1}.
\end{equation}
We find that the simple frequency coding  outperforms hash grid encoding~\cite{muller2022instant} which tends to generate surfaces with significant noise. To maintain temporal consistency, we follow~\cite{Ren2021PIRendererCP,Ma2023OTAvatarOT} to represent the deformation of any timestamp by the window of adjacent 27 frames of expression coefficients, which are averaged into a 256-dimensional feature vector by a trainable convolutional encoder.

\subsection{Neural rendering}\label{sec:Neural rendering}
This procedure aims to automatically texture the triangular meshes $(\mathbf{V}_{tri}, \mathbf{T}_{tri})$ extracted by \textit{MT} algorithm, and generate photorealistic images $ \hat{\mathbf{I}}$ given camera pose $\mathbf{P}$ and talking signal $\boldsymbol{\alpha}$. In the realm of deformable reconstruction~\cite{Newcombe_Fox_Seitz_2015,Lombardi_2019,Park2020NerfiesDN,Lombardi_2019}, the canonical or template model serves an important role in reducing the complexity of texture parametrization. Thanks to the well-designed geometry model in DynTet, defining a canonical space becomes straightforward. We upgrade the vanilla rendering process by incorporating a canonical projection, thereby unifying both shading and lighting within a single space.

\vspace{-1em}
\paragraph{Canonical projection.}
We propose the canonical projection $\mathrm{CanProj}$ to map the pixel coordinates $\mathcal{C}_{2d}$ from an arbitrary camera pose $\mathbf{P}$ into a canonical 3D space where the mean  expression lies.  This process is expressed as
\begin{equation}
	\begin{aligned}
		& \bar{\mathcal{C}}_{3d} = \mathrm{CanProj}(\mathcal{C}_{2d}, \mathbf{V}_{tri}, \mathbf{P}; \bar{\mathbf{V}}_{tri}, \mathbf{T}_{tri}), \\ 
		& \text{where}\ \   \bar{\boldsymbol{\alpha}} = \Sigma_{i=1}^{n} \boldsymbol{\alpha}_i / n, \\
		& \text{and} \ \ \ \  \bar{\mathbf{V}}_{tri}, \mathbf{T}_{tri} = \mathrm{MatTet}(\boldsymbol{F}(\mathbf{V}_{tet},\bar{\boldsymbol{\alpha}})).
	\end{aligned}
\end{equation}
In practical implementation, $\mathrm{CanProj}$ can be executed readily using a modern mesh rasterizer to derive coordinates $\mathcal{C}_{3d}$ and 2D barycentric coordinates within $(\mathbf{V}_{tri}, \mathbf{T}_{tri})$. Then, $\bar{\mathcal{C}}_{3d}$ positioned  in the canonical space, is computed through barycentric interpolation over $(\bar{\mathbf{V}}_{tri},\mathbf{T}_{tri})$.

\vspace{-1em}
\paragraph{Appearance model.}
We follow previous work~\cite{Munkberg2021ExtractingT3,Burley2012PhysicallyBasedSA} to adopt a physically-base material model, which allows easy integration of 3D assets within existing engines. To parameterize the appearance mapping $\boldsymbol{G}$, we employ a coordinate-based neural network comprising a hash grid encoder~\cite{muller2022instant} for querying spatial features and a lightweight MLP for predicting materials. This network  takes the projected coordinates and talking signal as inputs:
\begin{equation}
	\mathcal{K}_{d}, \mathcal{K}_{orm}  = \boldsymbol{G}(\bar{\mathcal{C}}_{3d},\boldsymbol{\alpha}).
\end{equation}
Here, $\mathcal{K}_{d}$ comprises three-channel diffuse albedo, and $\mathcal{K}_{orm}$ is the set of occupancy $k_o$, roughness $k_r$, and metalness factors $k_m$ for the GGX normal distribution function~\cite{Cook_Torrance_1981}. Unlike prior works, we do not predict normals, but instead compute them directly using the triangular mesh in the model space, which yields similar results.
\vspace{-1em}
\paragraph{Lighting model.}
We leverage the image based lighting model, where the environment light is given by a trainable mapping $\boldsymbol{L}$.  For each  position $\mathbf{v} \in \mathcal{C}_{3d}$ with a normal vector $\mathbf{n}$, the color along direction $\boldsymbol{\omega}_{o}$ is computed by
\begin{equation}
	\begin{aligned}
		L(\mathbf{v}, \boldsymbol{\omega}_{o})= L_d(\mathbf{v})+L_s(\mathbf{v}, \boldsymbol{\omega}_{o}),  
	\end{aligned}
\end{equation}
where $L_d$ is the diffuse intensity and $L_s$ is the specular intensity. Consider a hemisphere with the incident direction $\Omega=\left\{\boldsymbol{\omega}_i: \boldsymbol{\omega}^{T}_{i} \mathbf{n} \geq 0\right\}$, the first diffuse term is computed by
\begin{equation}
	\begin{aligned}
		L_d(\mathbf{v})= (1-k_m) \boldsymbol{k}_d     \int_{\Omega} \small{\boldsymbol{L}_i}\left(\mathbf{v}, \boldsymbol{\omega}_i\right) (\boldsymbol{\omega}^{T}_i \mathbf{n}) \mathrm{d} \boldsymbol{\omega}_i.
	\end{aligned}
\end{equation}
And the second term is based on Cook-Torrance microfacet specular shading mode model~\cite{Cook_Torrance_1981} $r(\boldsymbol{\omega}_{o},\boldsymbol{\omega}_i)$
to compute
\begin{equation}
	\begin{aligned}
		& L_s(\mathbf{v}, \boldsymbol{\omega}_{o})=\boldsymbol{k}_s \int_{\Omega} r(\boldsymbol{\omega}_{o},\boldsymbol{\omega}_i) \small{\boldsymbol{L}_i}(\mathbf{v},  \boldsymbol{\omega}_i) (\boldsymbol{\omega}^{T}_i \mathbf{n}) \mathrm{d} \boldsymbol{\omega}_i, 
	\end{aligned}
\end{equation}
where the specular color $\boldsymbol{k}_s = (1 - {k}_m) \cdot 0.04  + {k}_m\boldsymbol{k}_d$.
Following~\cite{Munkberg2021ExtractingT3}, we parameterize $\boldsymbol{L}$ with a cube map with resolution $6 \times 512 \times 512$, and the calculate the hemisphere integration by the split-sum method~\cite{Hill2017PhysicallyBS}.  By aggregating the rendered pixel colors along the camera pose, we obtain the rendered image $\hat{\mathbf{I}} = \{k_o\cdot L(\mathbf{v}, \boldsymbol{\omega}_{o})| \mathbf{v} \in \mathcal{C}_{3d})\}$.

\subsection{Training losses}\label{sec:Loss function}
Thanks to its hybrid representation, DynTet possesses an interpretable SDF space and a cost-efficient rendering procedure. This unique combination allows us to supervise its training in both image and geometry space, facilitating the estimation of reasonable 3D models from frontal images. 

\vspace{-1em}
\paragraph{Image supervision.} Given an image $\mathbf{I}$ and its prediction $\hat{\mathbf{I}}$, we train DynTet with the MSE loss for pixel-level reconstruction, and apply the overall LPIPS loss~\cite{zhang2018lpips} for enhancing sharp details~\cite{Li2023EfficientRN}. Besides, we find silhouette loss is important to provide shape guidance. The image loss is 
\begin{equation}
		\mathcal{L}_{img} = \mathcal{L}_{mse}(\hat{\mathbf{I}}, \mathbf{I}) + \mathcal{L}_{mse}(\hat{\mathbf{I}}_o, \mathbf{I}_o) + \lambda_{1}\small{\mathcal{L}_{LPIPS}}(\hat{\mathbf{I}}, \mathbf{I}),		
\end{equation}
where $\hat{\mathbf{I}}_o$ and $\mathbf{I}_o$ denote the binary masks of  head regions. We combine the head, background and torso together to train in practice to prevent noise around the facial contours.

\paragraph{3DMM supervision.} 
Frontal head views in talking videos often lack adequate depth information, resulting in flawed geometry estimation and artifacts in synthesized profile face. To counter this, we leverage 3DMM~\cite{Blanz1999AMM} as a geometry prior to mitigate these issues in the SDF space. We introduce two key losses to incorporate 3DMM. First, the normal distance loss $\mathcal{L}_{ndl}$ constrains the canonical model:
\begin{equation}
	\mathcal{L}_{ndl} = \mathbb{E}_{s\sim U(-a,a)} \|f_1(\mathbf{V}^{\bar{\boldsymbol{\alpha}}}_{3dmm} + s\cdot \small{\mathbf{N}}^{\bar{\boldsymbol{\alpha}}}_{3dmm}) - s\|^2_2,
\end{equation}
 where $\mathbf{V}^{\bar{\boldsymbol{\alpha}}}_{3dmm}$ and $\mathbf{N}^{\bar{\boldsymbol{\alpha}}}_{3dmm}$ represent 3DMM vertices and normal vectors generated by coefficient $\bar{\boldsymbol{\alpha}}$, while $a$ is a small value which we set to $0.1$. $\mathcal{L}_{ndl}$ injects facial prior into $f_1$, affecting the SDF values around 3DMM surface. Consequently, it avoids excessive restraint compared to full-space supervision~\cite{Park2019DeepSDFLC}, enabling adaptive learning in other regions. The second facial deformation loss is expressed as
\begin{equation}
	\mathcal{L}_{fdl} = \|[f_{1}(\mathbf{V}_{3dmm})]_{e}f_2(\small{\mathbf{V}_{3dmm}},\boldsymbol{\alpha}) - \mathbf{U}_{exp}(\boldsymbol{\alpha} - \bar{\boldsymbol{\alpha}}) \|^2_2.
\end{equation}
This loss helps the tetrahedral grid to mimick the 3DMM deformation. Although these losses target 3DMM vertices, the tetrahedral vertices are appropriately constrained due to their spatial density covering the 3DMM vertices.

\begin{figure*}[h]
    \centering
    \includegraphics[width=1\linewidth]{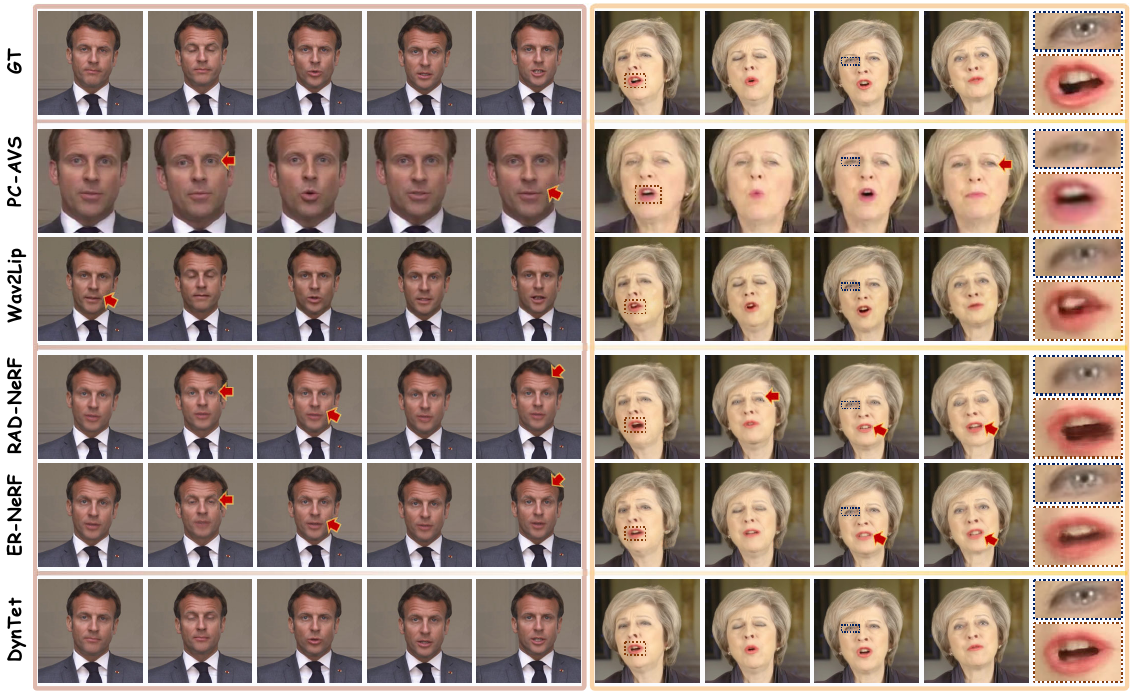}
    \caption{Qualitative comparison of DynTet with the prior methods~\cite{zhou2021pcavs,prajwal2020wav2lip,tang2022rad,Li2023EfficientRN}. Some representative defects are marked with red arrows, around which the generated eyes, mouths or wrinkles highlight discrepancies with real ones. The right panel presents the details of the mouth and eye area. The results show the superior realism and motion accuracy achieved by DynTet compared to existing methods.}
    \label{fig:compare1}
    \vspace{-3mm}
\end{figure*}

\paragraph{Loss function.}  We introduce a regularization term for the elastic score, denoted as $\mathcal{R}_{es} = \| [f_{1}(\mathbf{V}_{3dmm})]_{e} \|^2_2$, to reinforce the rigid property. We incorporate the weighted combination of the above constraints as follows to formulate the loss function described in Eq.~\eqref{eq:target}
\begin{equation}
    Loss = \mathcal{L}_{img} + \lambda_2\mathcal{L}_{ndl} + \lambda_3\mathcal{L}_{fdl} + \lambda_4\mathcal{R}_{es}.
\end{equation}

\section{Experiment}
\subsection{Experimental settings}

\paragraph{Dataset and pre-processing.} For a fair comparison, we follow recent works~\cite{guo2021ad,shen2022dfrf,Li2023EfficientRN} to conduct experiments on a publicly-released video dataset, which includes four high-definition talking videos with an average length of about 6500 frames at 25 FPS. Each raw video is cropped and resized to $512 \times 512$, except for the Obama data with the resolution $448 \times 448$. Each video is divided into training and test sets at a ratio of 10:1. We ensure strict alignment with the pre-processing steps outlined in AD-NeRF~\cite{guo2021ad}. In order to extract the 3DMM coefficients efficiently, we utilize a pre-trained model from Deng \textit{et al.} \cite{Deng2019Accurate3F}.

\vspace{-1em}
\paragraph{Implementation details.} We implement DynTet based on the code of \text{Nvdiffrec}\footnote{\url{https://github.com/NVlabs/nvdiffrec}}~\cite{Munkberg2021ExtractingT3} for differentiable Marching Tetrahedra and rasterization. We use a tetrahedral grid with a resolution of $128^3$~\cite{Doran2013IsosurfaceSI}, which is a commonly used setting \cite{Munkberg2021ExtractingT3,Gao2022GET3DAG}.
For the MLPs, we equip each one with ReLU-Linear layers. 
The geometry model utilizes a frequency positional encoder \cite{Tancik2020FourierFL} with $L=6$ and a 3-layer MLP with 128 middle neurons. Additionally, a 4-layer convolutional encoder \cite{Ren2021PIRendererCP} is used for the 3DMM coefficients. The appearance model consists of a 5-layer MLP with 256 middle neurons. During training, we perform 20,000 iterations with a batch size of 4. We employ the Adam optimizer with an initial learning rate of $1\times 10^{-3}$, which exponentially decayed to $1\times 10^{-5}$. The default hyperparameters are $\lambda_{1} = 0.1$, $\lambda_{2} = 100$, $\lambda_{3} = 100$, and $\lambda_{4} = 100$. All experiments are conducted on a single NVIDIA Tesla V100. 

\vspace{-2em}
\paragraph{Comparison baselines.}
We compare our method with several recent representative one-shot and person-specific models, including Wav2Lip~\cite{prajwal2020wav2lip}, PC-AVS~\cite{zhou2021pcavs}, NVP~\cite{thies2020nvp}, SynObama~\cite{suwajanakorn2017synthesizing}, SadTalker~\cite{Zhang2022SadTalkerLR}. In addition, we also compare our method with three end-to-end NeRF-based models: ADNeRF~\cite{guo2021ad}, RAD-NeRF~\cite{tang2022rad} and ER-NeRF~\cite{Li2023EfficientRN}. All these methods are implemented with their official code. 

\begin{table*}[ht]
\resizebox{1\linewidth}{!}{
        \setlength{\tabcolsep}{3.7mm}
        \centering
        \begin{tabular}{l|cccc|cccc|ccc}
        \hline\hline
        Methods & PSNR $\uparrow$ & LPIPS $\downarrow$ & FID $\downarrow$ & CSIM $\uparrow$ & LMD$^{m}$ $\downarrow$ & LMD$^{e}$ $\downarrow$ & AUE $\downarrow$ & Sync $\uparrow$ & Time & FPS  \\ \hline
        Ground Truth  & $\infty$            & 0               & 0              & 1.000    & 0          & 0    & 0          & 7.899          & -   & -          \\ \hline
        Wav2Lip~\cite{prajwal2020wav2lip}      & 31.15  & 0.0730            & 20.70          & 0.970   & 3.072  & 2.147     & 2.059 & \uline{8.256} & -   & 20    \\
        PC-AVS~\cite{zhou2021pcavs}       & 22.06          & 0.1345          & 47.53         & 0.802  & 2.496     & 3.609    & 4.283          & \textbf{8.540}          & -   & \uline{32}   \\
        AD-NeRF~\cite{guo2021ad}      & 30.41          & 0.0799          & 14.92          & 0.926 & 4.317  & 2.405       & 4.374          & 5.015          & 18h & 0.08       \\
        RAD-NeRF~\cite{tang2022rad}     & 31.51    & 0.0675          & 11.14          & 0.951 & 3.006 &2.285         & 3.317          & 4.409          & 5h  & 23         \\
        ER-NeRF{\small\dag}~\cite{Li2023EfficientRN}  & 32.50          & \uline{0.0345} & \uline{6.44} & \text{0.960} & \text{2.924} & \text{2.220} & \text{2.773}    & 4.944          & \textbf{2h}  & \text{23}        \\ \hline
        DynTet   & \textbf{35.15}         & 0.0619 & 12.23 & \uline{0.975} & \uline{2.418} & \uline{2.137} &  \uline{1.833}    & 7.407          & \textbf{2h}  & \textbf{46}        \\ 
        
        DynTet{\small\dag}  & \uline{34.84}         & \textbf{0.0223} & \textbf{4.72} & \textbf{0.978} & \textbf{2.284} & \textbf{2.129} & \textbf{1.712}    & 7.646          & \uline{3h}  & \textbf{46}        \\ \hline\hline
        \multicolumn{6}{l}{\small\dag \  supervised by overall LPIPS~\cite{zhang2018lpips}. }
        \end{tabular}
    }
    \setlength{\abovecaptionskip}{0cm}
    \caption{Quantitative results of the self-driven head reconstruction. The best and second results are in \textbf{bold} and \uline{underline}. 
    Wav2Lip takes ground truth as input, thus its PSNR and LPIPS values are biased. The FPS values are tested on the resolution of $512\times 512$.}
    \label{tab:setting1}
    \vspace{-0.4cm}
\end{table*}

\begin{figure}[t]
    \centering
    \includegraphics[width=1\linewidth]{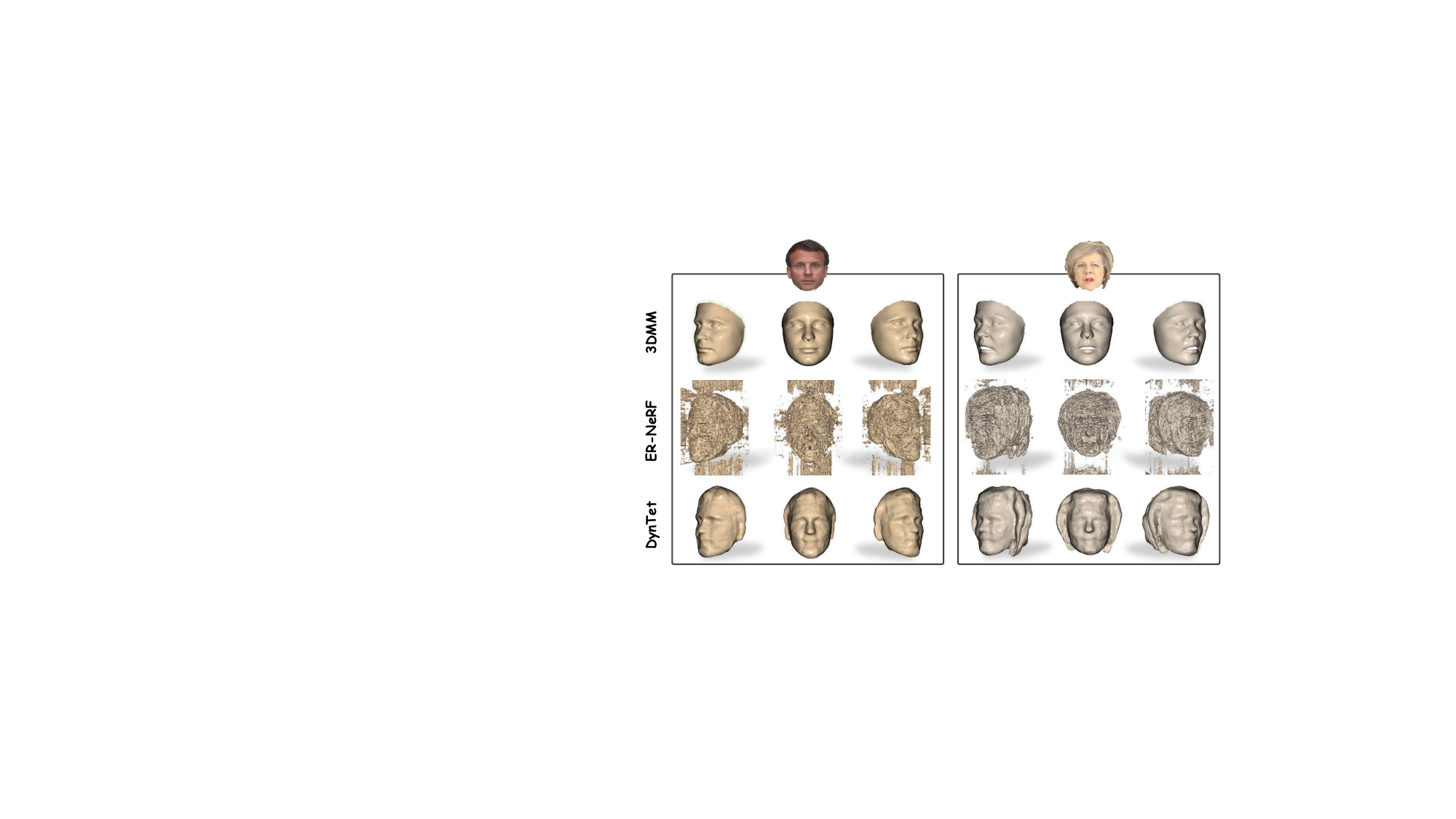}
    \caption{The triangular meshes from 3DMM~\cite{Blanz1999AMM}, ER-NeRF~\cite{Li2023EfficientRN} and DynTet. The surfaces extracted from ER-NeRF using the Marching Cubes~\cite{Lorensen1987MarchingCA} exhibit noise and undesirable topology. Note that the backs of the meshes may have some defects due to limited training data, but it does not impact the rendering results.}
    \label{fig:comparemesh}
    \vspace{-3mm}
\end{figure}

\subsection{Quantitative results}

\paragraph{Comparison settings.} We follow recent works to organize our comparisons into two settings: 1) Self-driven head reconstruction setting, in which we train a model for each video clip and evaluate the reconstruction quality on its respective test set.  2) Cross-driven lip synchronization setting, where we use clips from unseen videos to drive all methods for comparisons in lip synchronization. We extract two video clips from the public demos of SynObama \cite{suwajanakorn2017synthesizing} and NVP \cite{thies2020nvp}, which we refer to as Testset A and Testset B. 

\vspace{-1em}
\paragraph{Evaluation metrics.}  We assess the methods with several metrics: We evaluate the reconstruction quality using PSNR and LPIPS metrics~\cite{zhang2018lpips}. Realism and identity preservation are measured using Fréchet Inception Distance~(FID)~\cite{Heusel2017GANsTB} and Cosine Similarity of Identity Embedding~(CSIM) \cite{Ma2023OTAvatarOT}. Driving accuracy is assessed by calculating the landmark distances for the mouth ($\text{LMD}^{m}$) and eyes ($\text{LMD}^{e}$)~\cite{Chen2018LipMG}, while face motion accuracy is quantified by action units error~(AUE) \cite{baltrusaitis2018openface}. Temporal lip synchronization is evaluated using the SyncNet confidence score~(Sync) \cite{chung2017syncnet1}. Additionally, we report training time and frame-per-second (FPS) to evaluate the running efficiency of the methods. 

\vspace{-1em}
\paragraph{Head reconstruction.}
The results of the head reconstruction setting are presented in Table~\ref{tab:setting1}. For fairness, we provide DynTet  with and without LPIPS supervision, both of which demonstrate significant advantages in reconstructing accurate details and precisely controlling facial movements. While 2D-based methods such as Wav2Lip~\cite{prajwal2020wav2lip} and PC-AVS~\cite{zhou2021pcavs} excel in the lip synchronization due to the pre-training on the large-scale dataset, they fall short in the faithful appearance. Among NeRF-based methods, the recent work ER-NeRF~\cite{Li2023EfficientRN} achieves top performance in the evaluation. As the first tetrahedra representation for talking heads, DynTet notably outperforms on all metrics compared to NeRF-based approaches, showcasing its advancements in faithful reconstruction and accurate mapping from conditions to facial deformation. This further validates the great potential of tetrahedra meshes for dynamic modeling.

\begin{table}[t]

\resizebox{1\linewidth}{!}{
\centering
\setlength{\tabcolsep}{3mm}
\begin{tabular}{l|cc|cc}
\hline\hline
   \multirow{2}{*}{Methods}         & \multicolumn{2}{c|}{Testset A} & \multicolumn{2}{c}{Testset B} \\ \cline{2-3}  \cline{4-5}  
      & AUE $\downarrow$   & Sync $\uparrow$  & AUE $\downarrow$   & Sync $\uparrow$ \\ \cline{1-5}  
Ground Truth & 0            & 7.386          & 0             & 6.676         \\ \hline
SynObama \cite{suwajanakorn2017synthesizing}     & 5.574        & 7.419          & -             & -             \\
NVP \cite{thies2020nvp}          & -            & -              & 7.954         & 6.562         \\
Wav2Lip \cite{prajwal2020wav2lip}      & 5.029        & \textbf{8.394} & {7.415}& \uline{9.072}         \\
PC-AVS \cite{zhou2021pcavs}       & 4.359       & \uline{8.087}          &  7.450        & \textbf{9.964}         \\
SadTalker \cite{Zhang2022SadTalkerLR}          & 4.732 & 7.207          &      6.760    & 7.932         \\ \hline
AD-NeRF \cite{guo2021ad}      & {4.277}        & 6.041          & {6.731}        & 5.567         \\
RAD-NeRF \cite{tang2022rad}     & \uline{4.172}        & 6.541          & {6.733}         & 6.786         \\ 
ER-NeRF~\cite{Li2023EfficientRN}         & 4.210        & {6.877}          & {6.669}         & {7.401}         \\ \hline
DynTet       & \textbf{3.672}        & {5.055}          & \textbf{6.029}         & {6.401}         \\
Audio2Exp + DynTet        & 4.316        & {7.055}          & \uline{6.541}         & {7.335}         \\ \hline\hline
\end{tabular}
 }
 
\caption{Quantitative results of cross-driven lip synchronization. The best and second results are in \textbf{bold} and {\ul underline}. To drive DynTet with audios, we utilize an off-the-shelf model Audio2Exp~\cite{Zhang2022SadTalkerLR} to convert audios into 3DMM coefficients. }
\label{tab:setting2}
\vspace{-1em}
\end{table}

\vspace{-1em}
\paragraph{Lip synchronization.}

DynTet offers both frame-driven and audio-driven approaches to accomplish this task. In the frame-driven approach, 3DMM coefficients are directly extracted from the target videos. On the other hand, the audio-driven approach utilizes an off-the-shelf model, specifically the Audio2Exp model in SadTalker~\cite{Zhang2022SadTalkerLR}, to convert audios into the 3DMM coefficients. It is worth noting that the target coefficients, denoted as $\boldsymbol{\alpha}'$, and the training coefficients $\boldsymbol{\alpha}$, may have different distributions according to speaking habits. We find that a re-center process, expressed as $\boldsymbol{\alpha}' + (\bar{\boldsymbol{\alpha}} -\bar{\boldsymbol{\alpha}}')$ is crucial to rectificate the talking style. As shown in Table~\ref{tab:setting2},  the frame-driven approach of  DynTet unsurprisingly achieves the best AUE performance among all methods. Furthermore, the audio-driven approach of DynTet also achieves comparable results with ER-NeRF and SadTalker, indicating its flexibility and strong generalization properties for various applications.

\paragraph{Running efficiency.}
Table~\ref{tab:setting1} shows that DynTet achieves fast training speed and real-time inference at $512\times 512$ resolution. Notably,  DynTet maintains the same 46 FPS even at $1024\times 1024$ resolution, thanks to the cost-efficient rasterization rendering. In contrast, ER-NeRF and RAD-NeRF experience a half drop in FPS. This highlights our efficiency advantage in challenging high-resolution scenes.

\begin{figure}[t]
    \centering
    \includegraphics[width=1\linewidth]{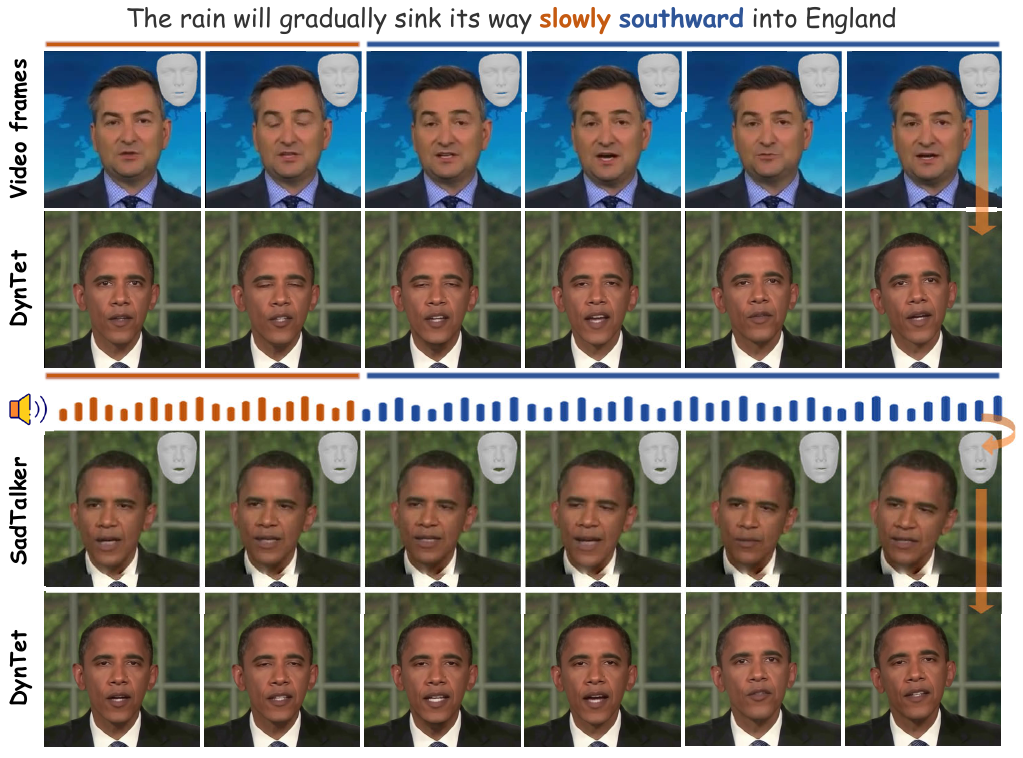}
    \vspace{-2em}
    \caption{Qualitative results of the cross-driven setting. The top and bottom panels show the frame- and audio-driven results, respectively. We attach the estimated 3DMM shapes for reference. }
    \label{fig:compare2}
    \vspace{-1em}
\end{figure}

\subsection{Qualitative results}
Figure~\ref{fig:compare1} demonstrates that DynTet effectively addresses challenges faced by prior methods. It generates photo-realistic images with intricate details in non-rigid areas and achieves precise control of mouth, blinks, and even wrinkles. Our \textit{supplementary videos} showcase the impressive temporal stability of DynTet, resulting in smooth and stable motion. In Figure~\ref{fig:comparemesh}, the comparison of meshes obtained by different methods reveals that DynTet outperforms 3DMM and NeRF in terms of topology. Additionally, Figure \ref{fig:compare2} showcases the accurate expression control achieved by frame-driven DynTet, while audio-driven DynTet surpasses SadTalker in producing realistic and expressive results. These advancements firmly establish DynTet as a promising approach for modeling talking heads.

\begin{figure}[t]
    \centering
    \includegraphics[width=1\linewidth]{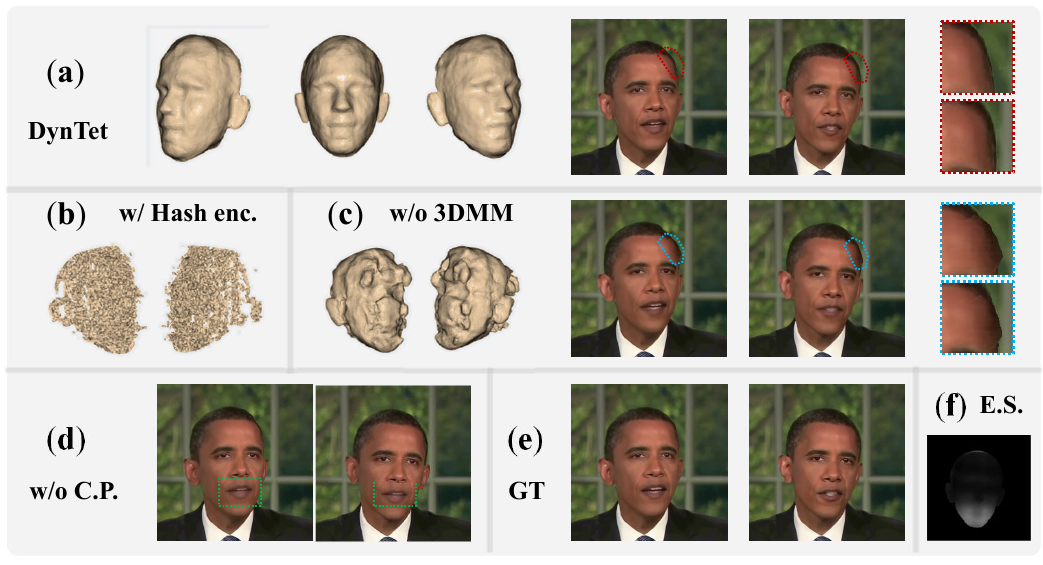}
    \vspace{-2em}
    \caption{The validation of key components in DynTet. (a) Result of DynTet. (b) Replacing frequency encoding with hash encoding~\cite{muller2022instant}.  (c) Removing 3DMM supervision. (d) Removing canonical projection. (e) Groundtruth. (f) Visualization of elastic scores.}
    \label{fig:ablation}
    \vspace{-1em}
\end{figure}

\subsection{Ablation study}
We present the results of different structures in Table~\ref{tab:ablation} and Figure~\ref{fig:ablation}. It is evident that the removal of canonical projection (C.P.) has a quite negative impact, resulting in increased LPIPS and LMD values. This is because the absence of the canonical projection increases the difficulty of texture learning and consequently hinders the supervision of loss on the geometry model. Similarly, removing 3DMM supervision leads to flawed meshes and introduces artifacts into the images. The visualization of elastic scores (E.S.) highlights its function in quantifying the non-rigid property, and its removal affects the temporal stability of the results (see the supplementary videos). Interestingly, we find
 replacing frequency encoding with hash encoding leads to a cluttered mesh, indicating that it is desirable to encode coordinates as low-frequency signals in geometry learning. In addition, we explore the impact of middle neuron channels within the appearance mapping, and find that dimensions ranging from 64 to 256 strike a balance between expressivity and inference speed. These findings underscore the reasonable designs of DynTet for achieving desirable results.

\begin{table}[t]
\resizebox{1\linewidth}{!}{
\centering
\small
\setlength{\tabcolsep}{1mm}
\begin{tabular}{c|c|ccc|cccc}
\hline\hline
      & DynTet & - 3DMM & - C.P. & - E.S.   & 32 & 64 & 128  & 512 \\ \hline
      FPS$\uparrow$   &  46 & - & - & - &  103  & 92 &   80     &  23  \\ 
\#Param & 1.41M & - & - & -  & 1.11M  &  1.13M  &     1.19M   & 2.28M   \\ \hline
PSNR$\uparrow$    & 34.84 & 33.65 & 34.37 & 34.81  &  32.22  &  34.20  &     35.15  & 35.06  \\ 
LPIPS$\downarrow$    & 0.0223 & 0.0686 & 0.0529  &  0.0247 &  0.0539  &  0.0254  &     0.0237  & 0.0219  \\
LMD$^{m}$ $\downarrow$    &  2.284 & 2.453 & 2.572 & 2.336  & 2.875  &  2.636  &   2.449  & 2.201       \\ 
\hline\hline
\end{tabular}
 }
\vspace{-0.5em}
\caption{Quantitative ablation of DynTet via removing 3DMM supervision, canonical projection, elastic score or changing the number of middle neurons in appearance mapping from $32$ to $512$. $\#$Param denotes the parameter number of DynTet.}
\label{tab:ablation}
\vspace{-1.5em}
\end{table}

\section{Conclusion} 

We introduce Dynamic Tetrahedra (DynTet), a novel hybrid representation for realistic and expressive talking heads. DynTet upgrades tetrahedral meshes from statics to dynamics with a new architecture, a canonical space and guidance from geometry prior.
DynTet can efficiently generate high-resolution talking videos with realism and precise motion control beyond prior works. Our work may inspire future research in the direction of Dynamic Tetrahedra. 

\noindent\textbf{Acknowledgements.}
This paper is supported by the Strategic Priority Research Program of Chinese
Academy of Sciences (XDA27000000), the National key research and development program of China (2021YFA1000403), the National Natural Science Foundation of China (Nos. U23B2012, 11991022) and the Fundamental Research Funds for the Central Universities  (E3E41904).

{
    \small
    \bibliographystyle{ieeenat_fullname}
    \bibliography{main}
}

\clearpage
\setcounter{page}{1}
\maketitlesupplementary

\section{Limitations} As an original work, Dyntet does face several challenges, such as effectively representing fine-grained textures like hairs and handling the driving of large deformations. These warrant the need for further research.

\section{More Ablation}
\begin{figure}[htbp!]
    \centering
\includegraphics[width=1\linewidth]{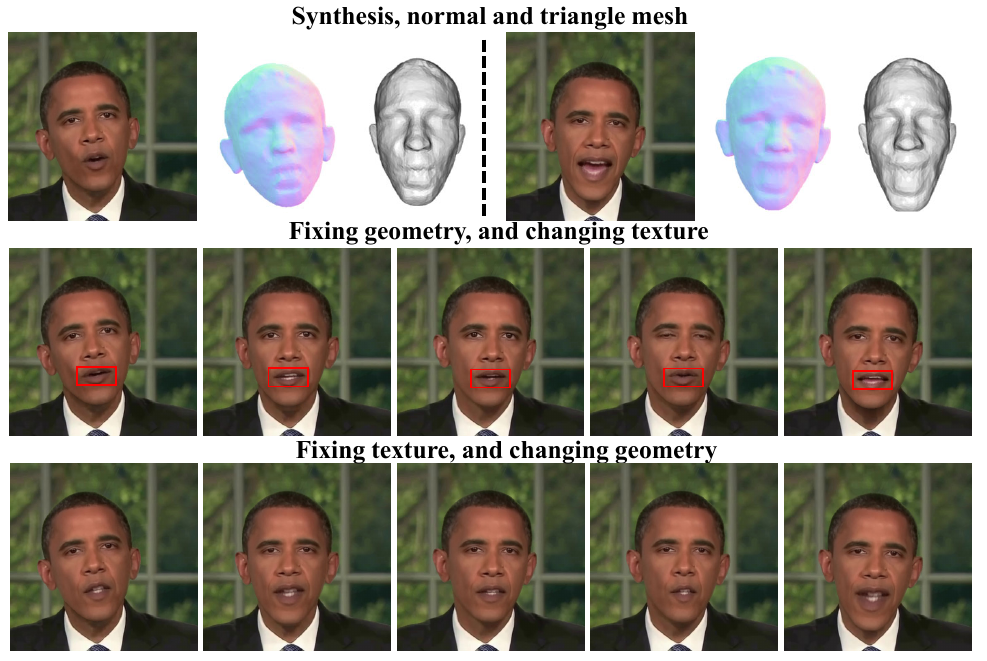}
    \caption{Effects of dynamic texture and geometry.}
    \label{fig:ablation 2}
    \vspace{-2em}
\end{figure}
\paragraph{Analyzing effects of dynamic texture and geometry on mouth movement.} 
We provide a comprehensive analysis in Fig.~\ref{fig:ablation 2}. From the meshes and images, it is evident that the geometry dictates the mouth boundary and size, while the dynamic texture primarily enhances the intricate details of the inner mouth, such as teeth and tongue. This design addresses the limitations of triangle meshes in representing complex topology and fine details.  

\begin{figure}[htbp!]
    \centering
\includegraphics[width=1\linewidth]{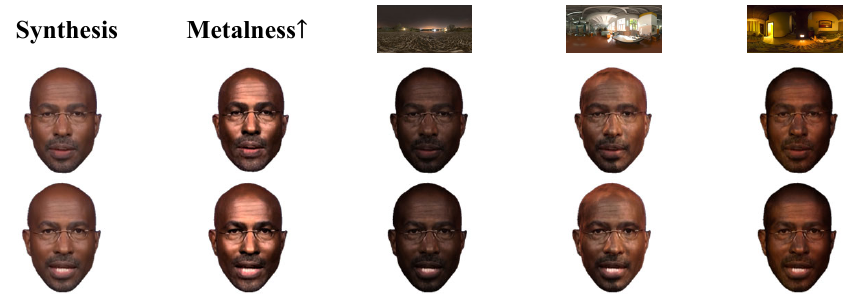}
    \caption{Editing material metalness, and relighting heads.}
    \label{fig:relight}
    \vspace{-2em}
\end{figure}

\paragraph{Analyzing the material export and relighting abilities.} 
Thanks to the utilization of PBR materials and lighting models, editing talking heads becomes straightforward. As shown in Fig.~\ref{fig:relight}, we can easily impart a metallic luster to heads by adding a positive value to the exported metalness factors, and relight heads using arbitrary HDRi maps.

\begin{figure}[htbp!]
    \centering
\includegraphics[width=1\linewidth]{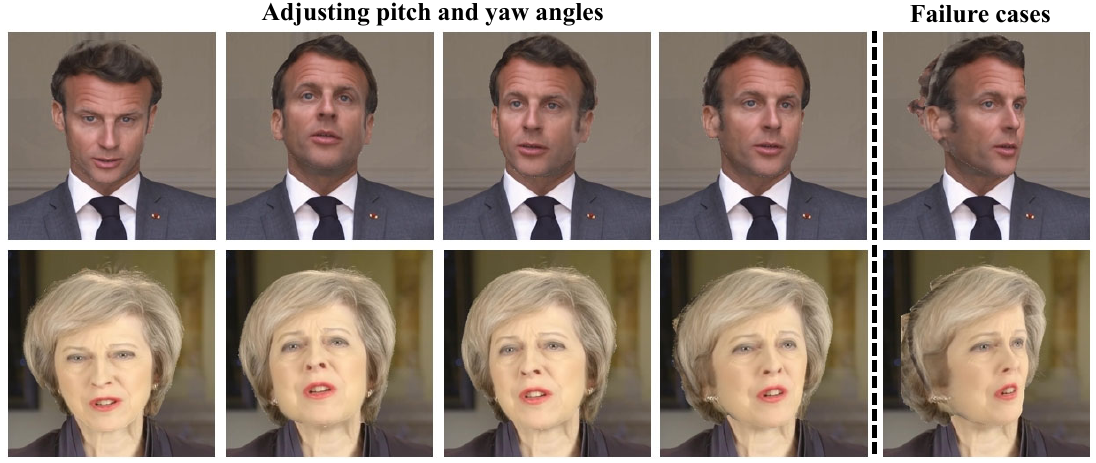}
    \caption{Novel view synthesis with poses beyond dataset.}
    \label{fig:novelview}
    \vspace{-1em}
\end{figure}

\paragraph{Analyzing the capability of novel view synthesis.} 
DynTet exhibits nice performance on this task. As depicted in Fig.~\ref{fig:novelview}, by first identifying the extreme Euler angles in the dataset, DynTet can generate realistic and consistent heads even when angles are beyond them. However, as the training data primarily includes frontal faces, artifacts may arise when generating the back of the head.

\section{Supplementary Video} We provide comprehensive video results. Please refer to \url{https://youtu.be/Hahv5jy2w_E}.


\end{document}